\documentclass{paper}  %
\usepackage[preprint, nonatbib]{neurips_2026}
\usepackage[numbers]{natbib}

\usepackage[utf8]{inputenc} 
\usepackage[T1]{fontenc}    
\usepackage{hyperref}       
\usepackage{url}            
\usepackage{booktabs}       
\usepackage{amsfonts}       
\usepackage{nicefrac}       
\usepackage{microtype}      
\usepackage{xcolor}         
\usepackage{amsmath,amsfonts,amssymb,amsthm}
\usepackage{algorithm}
\usepackage{algpseudocode}
\usepackage{graphicx}
\usepackage{caption}
\usepackage{subcaption}
\usepackage{xcolor}
\usepackage{mathtools}
\usepackage{pstricks}
\usepackage{tikz}
\usepackage{enumitem}

\DeclareMathOperator*{\argmin}{arg\,min}

\DeclarePairedDelimiterX{\infdivx}[2]{(}{)}{%
  #1\;\delimsize\|\;#2%
}
\newcommand{\infdiv}{\text{KL}\infdivx}

\newtheorem{definition}{Definition}

\newtheorem{lemma}{Lemma}
\newtheorem{theorem}{Theorem}

\newtheorem{corollary}{Corollary}
\newtheorem{remark}{Remark}
\newtheorem{proposition}{Proposition}

\newcommand{\Tref}{T^{\mathrm{ref}}_k}
\newcommand{\Tcon}{T^{\mathrm{proj}}_k}

\title{\LARGE \bf
Drift-admissibility via Optimal Transport
}

\author{%
  Abigail Langbridge\\
  Dyson School of Design Engineering\\
  Imperial College London\\
  \texttt{abigail.langbridge18@imperial.ac.uk} \\
  \And
  Enrico Biffis\\
  Imperial Business School\\
  Imperial College London\\
  \texttt{e.biffis@imperial.ac.uk}\\
  \And
  Jakub Mare\v{c}ek\\
  Department of Computer Science\\
  Czech Technical University in Prague\\
  \texttt{jakub.marecek@fel.cvut.cz}\\
  \And
  Robert Shorten\\
  Dyson School of Design Engineering\\
  Imperial College London\\
  \texttt{r.shorten@imperial.ac.uk}\\
}

\begin{document}

\maketitle
\thispagestyle{empty}
\pagestyle{empty}

\begin{abstract}
We introduce structure-aware optimal transport diagnostics for assessing
whether empirical time-series dynamics are compatible with a prescribed
predictable-drift class. Our approach translates a pathwise property of
stochastic processes into a finite-dimensional projection problem over
empirical transition laws, encoding admissibility as row-wise moment
constraints on local couplings and quantifying deviations via a
Kullback--Leibler projection gap $G_{\lambda,k}$. This yields both a
principled diagnostic and a scalable Bregman--Dykstra algorithm that
retains the efficiency of Sinkhorn-type methods while incorporating
structural constraints. Calibrated against a Brownian reference null,
$G_{\lambda,k}$ separates process classes that are indistinguishable by
second-order statistics alone---including fractional Brownian motion and
BSS processes sharing identical Hurst indices---without asymptotic
resampling or continuous-time recovery. This positions the method as a
general tool for structure-aware model validation of sequential data.
\end{abstract}

\section{Introduction}
A central challenge in modelling time-series data is determining
whether observed dynamics are compatible with a given structural
hypothesis.
In stochastic analysis, semimartingale structures are a key
hypothesis, underpinning arbitrage-free modelling, stochastic
integration, and continuous-time financial theory
\cite{JacodShiryaev,BeiglbockSchachermayerVeliyev2011,Rogers1997}.
However, verifying such a structure from discrete observations is
notoriously difficult: classical characterisations are pathwise or
asymptotic and do not translate directly into finite-sample,
data-driven diagnostics
\cite{JacodProtter,barndorff2004power,ait2009estimating,bibinger2024jump}.
Optimal transport (OT) provides scalable tools for learning couplings,
but standard formulations typically constrain only marginals and costs
rather than conditional structure.

In this work, we bridge these two perspectives.
We introduce a new class of \emph{constraint-aware OT projections}
that encode predictable-drift admissibility as row-wise moment
constraints on empirical transition laws.
The term \emph{structure-aware} refers precisely to this: the
constraints act on the \emph{conditional} (row-wise) distribution of
increments given the predictor state, rather than on global marginals
or aggregate costs as in standard Sinkhorn formulations.
This construction provides a finite-dimensional, observation-driven
diagnostic for a bounded predictable drift class inspired by special
semimartingales, replacing pathwise conditions with a convex
feasibility problem over transport plans. The resulting \emph{projection gap} $G_{\lambda,k}$ quantifies the
minimal KL deformation required to enforce the chosen structural
constraints; it is a compatibility diagnostic, not a standalone
statistical test (Section~\ref{sec:discussion}).



In this paper, we make the following four key contributions:
\begin{enumerate}[label=(\alph*)]
  \item We introduce a finite-dimensional optimal transport projection that encodes predictable-drift admissibility via row-wise moment constraints on empirical transition laws.
  \item We develop an efficient solver based on entropic OT and Bregman--Dykstra updates, combining Sinkhorn scaling with closed-form exponential tilting, validated against a \texttt{cvxpy}/MOSEK baseline (Section~\ref{sec:exp:solver}, Appendix~\ref{app:complexity}).
  \item We define the projection gap $G_{\lambda,k}$, calibrated against a Brownian reference null at a $62\%$ violation probability, and establish a monotonicity property (Proposition~\ref{prop:monotone}): $G_{\lambda,k}$ is non-increasing as the admissible class is enlarged.
  \item We use simulated processes---including Brownian motion, continuous It\^{o} diffusions, jump processes, fractional Brownian motion (fBM), and BSS-type processes---to demonstrate that $G_{\lambda,k}$ exhibits distinct scaling behaviour across information structures, capturing conditional drift compatibility beyond second-order statistics.
\end{enumerate}

\subsection{Related Work}
\label{sec:related}

\paragraph{Constrained Optimal Transport.}
Entropy-regularised OT \cite{cuturi2013advances}, introduced to accelerate
computation of Wasserstein distances via Sinkhorn iterations, admits a
natural extension to multiple convex constraints through iterative
Bregman projections \cite{benamou2015iterative}.
Several recent works have extended OT beyond the classical marginal
constraints, including unbalanced settings \cite{chizat2018scaling},
capacity constraints \cite{korman2015optimal}, and moment- or
feature-constraints \cite{blondel2018smooth,buyl2022optimal,tang2024sinkhorn}.
These approaches impose constraints at the level of global marginals
or aggregate statistics.
A related line of work studies OT under \emph{flow-rate constraints},
where aggregate mass through a constriction set is bounded above
\cite{dong2024monge}.
In contrast, we introduce row-wise (conditional) moment constraints
acting directly on the coupling, encoding structure at the level of
local transition laws.
This distinction is fundamental: flow-rate constraints bound aggregate
mass through a bottleneck (a global constraint), whereas our
$\mathcal{C}_+, \mathcal{C}_-$ constrain the conditional first moment
\emph{within each row individually}.
Using these constraints, we define a novel structure-aware divergence
as the KL projection gap to the nearest coupling satisfying
admissibility constraints.

\paragraph{Semimartingality and Discrete Diagnostics.}
In mathematical finance, semimartingales are the canonical class for
stochastic integration and self-financing trading strategies, and
non-semimartingale price models may lead to arbitrage pathologies
\cite{BeiglbockSchachermayerVeliyev2011,Rogers1997}.
This structural role is distinct from martingality, which is usually
imposed on discounted prices under an equivalent pricing measure
\cite{Delbaen1992,DelbaenSchachermayer1994}.
In applications, the process is not observed continuously: one
observes a finite grid of prices and therefore cannot recover the
latent continuous-time decomposition directly \cite{JacodProtter}.
Empirical diagnostics in this setting typically rely on high-frequency
asymptotics \cite{barndorff2004power,ait2009estimating,bibinger2024jump}.
These approaches test specific path properties but do not directly
assess the admissibility of predictable drift.
We instead adopt a finite-sample, distributional perspective,
replacing pathwise conditions with a convex projection problem on empirical transition laws.

\subsection{Observation Model}
\label{sec:benchmark}

Throughout, martingale and semimartingale properties are understood
relative to a fixed stochastic basis $(\Omega, \mathcal{F},
\mathbb{F}, \mathbb{P})$, where $\mathbb{F} :=
(\mathcal{F}_t)_{t\in[0,T]}$ is an increasing, right-continuous
filtration.
A process $(X_t)_{t \geq 0}$ is said to be $\mathbb{F}$-\emph{adapted}
if $X_t$ is $\mathcal{F}_t$-measurable for every $t \in [0,T]$
(written $X_t \in \mathcal{F}_t$); we refer to
\cite{JacodShiryaev} and Appendix~\ref{app:stochastic-analysis} for further
background.
A real-valued, c\`adl\`ag, $\mathbb{F}$-adapted process $X$ is an
$\mathbb{F}$-semimartingale if it admits a decomposition
\begin{equation*}
  X_t = X_0 + M_t + A_t, \quad t \in [0,T],
\end{equation*}
where $M$ is an $\mathbb{F}$-local martingale and $A$ is an
$\mathbb{F}$-adapted finite-variation process.
Here, we focus on the case of \emph{special semimartingales}, where
$A$ can be chosen to be predictable.

We fix a single observed grid $\pi = \{0 = t_0 < t_1 < \cdots <
t_N = T\}$ with mesh $\Delta t := \max_i(t_i - t_{i-1})$.
The asymptotic notation $\Delta t \to 0$ is used only to state the
structural question; the diagnostic itself is computed from this
single grid.
The data observed over $\pi$ are $X_{t_0}, X_{t_1}, \ldots,
X_{t_N}$, and the observable information is described by the
observation-based filtration $\mathbb{G} :=
(\mathcal{G}_i)_{i=0,\ldots,N}$, with $\mathcal{G}_i :=
\sigma(X_{t_0}, \ldots, X_{t_i})$.


The question studied here is not whether one can recover the latent decomposition $X = X_0 + M + A$ from the sampled path, but whether the observed one-step dynamics are compatible with the existence of such a decomposition when viewed through the observable filtration $\mathbb G$. Section~2 formalises this through predictable box edges.

\section{Drift-Admissibility Problem}
\label{sec:drift}

We now formulate a finite-dimensional diagnostic on the sampled
increments $\Delta_i X$.
At step $i$, a discrete predictable drift increment must be
$\mathcal{G}_{i-1}$-measurable.
We therefore prescribe lower and upper \emph{predictable box edges}
$d_{i,-}, d_{i,+} \in \mathcal{G}_{i-1}$, $d_{i,-} \leq d_{i,+}$
a.s., and interpret $[d_{i,-}, d_{i,+}]$ as the admissible range for
the predictable component of the one-step increment.

\begin{definition}[Predictable-box compatibility]
\label{def:compat}
Fix an observation grid $\pi$, the observed filtration $\mathbb{G}$,
and predictable box edges $(d_{i,-}, d_{i,+})_{i=1}^N$.
We say that the sampled increment system $(\Delta_i X)_{i=1}^N$ is
\emph{predictable-box compatible} with the box edges if, for every
$i = 1, \ldots, N$,
\begin{equation*}
  d_{i,-} \leq \mathbb{E}[\Delta_i X \mid \mathcal{G}_{i-1}] \leq
  d_{i,+} \quad \text{a.s.,}
\end{equation*}
whenever the conditional expectation is well defined.
\end{definition}

\begin{remark}[Doob-decomposition equivalence]
\label{rem:doob}
Predictable-box compatibility is equivalent to Doob decomposability
of the sampled increments with respect to $\mathbb{G}$, with the
additional requirement that the finite-variation part
$A_i = \mathbb{E}[\Delta_i X \mid \mathcal{G}_{i-1}]$ lies in
$[d_{i,-}, d_{i,+}]$ a.s.
It is deliberately \emph{weaker} than asserting that the underlying
continuous-time $X$ is a special semimartingale, which cannot be
certified from discrete data alone.
\end{remark}

Equivalently, compatibility means that there exist random variables
$(M_i, A_i)_{i=1}^N$ such that
\begin{equation*}
  \Delta_i X = M_i + A_i, \quad \mathbb{E}[M_i \mid
  \mathcal{G}_{i-1}] = 0, \quad A_i \in \mathcal{G}_{i-1},
\end{equation*}
with $d_{i,-} \leq A_i \leq d_{i,+}$ a.s.
A natural semimartingale-type specification is obtained by scaling
the box edges with the mesh, setting $d_{i,-} = \Delta t \,
\beta_{i,-}$, $d_{i,+} = \Delta t \, \beta_{i,+}$, with
$\beta_{i,-}, \beta_{i,+} \in \mathcal{G}_{i-1}$.
This scaling is consistent with It\^{o}-type models whose predictable
finite-variation component has locally bounded drift rate.

In finite samples, the conditional laws of $\Delta_i X$ are not
known.
The next section therefore replaces the conditional law of $\Delta_i
X$ given the observed past by a finite-dimensional coupling between
predictor states and increments.
The predictable box constraints then become linear row-wise
constraints on the conditional mean of the increment under that
coupling.

\section{Constraining One-Step Drift with Optimal Transport}
\label{sec:method}


We encode the observable past through a predictor variable $Z_i \in
\mathcal{G}_{i-1}$, $i = 1, \ldots, N$.
The predictor $Z_i$ may be finite-dimensional and may contain, for
example, the current observed state, lagged increments, realized
variation estimates, or other $\mathcal{G}_{i-1}$-measurable
summaries of the past.
In particular, the framework is not restricted to the Markovian
choice $Z_i = X_{t_{i-1}}$: a richer predictor encodes
non-Markovian dependence, while a low-dimensional predictor yields a
deliberately coarser diagnostic.
We represent the predictable box edges as measurable functions of the
predictor, $d_{i,-} = d_-(Z_i)$, $d_{i,+} = d_+(Z_i)$.

Let $Y_i := \Delta_i X$ denote the one-step increment.
For computational purposes we discretize the predictor space and the
increment space.
Let $\mu = \sum_{k=1}^m \mu_k \delta_{z_k}$ and $\nu =
\sum_{j=1}^\ell \nu_j \delta_{y_j}$ be discrete measures on
predictor and increment space, respectively.
A coupling $T = (T_{kj}) \in \mathbb{R}_+^{m \times \ell}$ between
$\mu$ and $\nu$ satisfies $T\mathbf{1} = \mu$,
$T^\top\mathbf{1} = \nu$.
The row indexed by $k$ corresponds to the conditional distribution of
the next increment given predictor value $z_k$.
The predictable box-edge constraints therefore become
\begin{equation*}
  \mu_k \, d_-(z_k) \;\leq\; \sum_{j=1}^\ell y_j \, T_{kj} \;\leq\;
  \mu_k \, d_+(z_k), \quad k = 1, \ldots, m.
\end{equation*}

\begin{remark}[Discretisation trade-off]
\label{rem:discretisation}
The predictor and increment spaces are discretised to make the
infinite-dimensional population problem tractable: a
continuous-state entropic-OT projection requires either a
parametric coupling family or discretisation, exactly as in
standard entropic OT~\cite{cuturi2013advances}.
The main cost of coarse discretisation is that predictor bins pool
states with different true conditional means, which can dilute
genuine drift-constraint violations.
However, as argued in Section~\ref{sec:gap:relax}, this dilution
is conservative---it suppresses $G_{\lambda,t}$, never inflates
it---and for processes with memory, the bin-averaged conditional
mean preserves the directional bias of the memory structure.
\end{remark}

More generally, one may replace the raw increment $y$ by a measurable
increment operator $g^n(z, y)$.
This allows for truncation, normalisation, or other
predictor-dependent transformations of the one-step increment.
The choice $g^n(z, y) = y_j - z_k$ yields the basic increment
formulation used throughout this paper.
This motivates the following feasible class of couplings:
\begin{align*}
  \mathcal{C}_n(d_-, d_+; g^n) := \Bigl\{T \in \mathbb{R}_+^{m
  \times \ell} : \:& T\mathbf{1} = \mu,\; T^\top\mathbf{1} = \nu,\; \\
  & \mu_k d_-(z_k) \leq \textstyle\sum_j g^n(z_k, y_j) T_{kj} \leq
  \mu_k d_+(z_k),\; k = 1, \ldots, m\Bigr\}.
\end{align*}

\subsection{Entropic Projection Formulation}
\label{sec:entropic}

Let $C \in \mathbb{R}^{m \times \ell}$ be a cost matrix.
In all experiments reported in this paper, $C_{k,ij} \equiv (Y_j -
Z_i)^2$, the matrix of pairwise squared distances between the
one-dimensional predictor and increment grids, acting as a smoothness
prior that favours cheap local transitions.
Define the strictly positive Gibbs kernel $K_0 := \exp(-C/\varepsilon)$
for $\varepsilon > 0$.
We use the generalised KL divergence $\infdiv{T}{K_0} :=
\sum_{k,j} \bigl(T_{kj} \log \frac{T_{kj}}{K_{0,kj}} - T_{kj} +
K_{0,kj}\bigr)$.
The hard-constrained entropic projection problem is
\begin{equation}
  \min_{T \in \mathcal{C}_n(d_-, d_+; g^n)} \infdiv{T}{K_0}.
  \label{eq:hard}
\end{equation}

\begin{remark}[KL divergence as information-theoretic deformation]
\label{rem:kl}
The choice of KL divergence is motivated by three complementary
considerations.
First, KL is the Bregman divergence generated by negative entropy, so
its minimiser over a linear moment constraint is always an
exponential-family member (Lemma~\ref{lem:tilt}), enabling
closed-form row-wise projections.
Second, it composes with Sinkhorn marginal scaling, so the entire
projection---marginals and drift constraints---is solved by multiplicative
updates (Lemma~\ref{lem:dykstra}).
Third, by Sanov's theorem, a KL projection gap is the exponential
rate at which the observed law is improbable under the nearest
admissible law; this is precisely why Hoeffding-type goodness-of-fit
tests are KL-tests.
\end{remark}

We isolate the marginal constraints $\mathcal{C}_\mu := \{T \geq 0 :
T\mathbf{1} = \mu\}$ and $\mathcal{C}_\nu := \{T \geq 0 :
T^\top\mathbf{1} = \nu\}$, and the drift-constraint halfspaces
\begin{align*}
  \mathcal{C}_+ &:= \Bigl\{T \geq 0 : \textstyle\sum_j g^n(z_k,
  y_j) T_{kj} \leq \mu_k d_+(z_k),\; k = 1, \ldots, m\Bigr\}, \\
  \mathcal{C}_- &:= \Bigl\{T \geq 0 : \textstyle\sum_j g^n(z_k,
  y_j) T_{kj} \geq \mu_k d_-(z_k),\; k = 1, \ldots, m\Bigr\}.
\end{align*}
The feasible set decomposes as $\mathcal{C}_n = \mathcal{C}_\mu \cap
\mathcal{C}_\nu \cap \mathcal{C}_+ \cap \mathcal{C}_-$.
This formulation makes explicit that predictable-drift admissibility
is a feasibility problem over conditional couplings.

\subsection{KL Projections with Predictable-Drift Constraints}

The sets $\mathcal{C}_+$ and $\mathcal{C}_-$ are convex but not
affine.
Standard Sinkhorn iterations therefore no longer guarantee
convergence; we consider the general projection problem
\begin{equation}
  \min_{T \in \bigcap_{l=1}^L \mathcal{C}_l} \infdiv{T}{K_0},
  \label{eq:general}
\end{equation}
where $\{\mathcal{C}_r\}_{r=1}^L$ denotes the collection of marginal
and drift-constraint sets.

\begin{lemma}[Dykstra's algorithm for KL projections
{\cite{bauschke2000dykstras}}]
\label{lem:dykstra}
Let $\{\mathcal{C}_l\}_{l=1}^L$ be closed convex sets with
non-empty intersection, and suppose that the KL projection is finite.
Then cyclic KL projections with Dykstra corrections converge to the
unique minimiser of the projection problem.
\end{lemma}

Since the row moment $\psi_k(\gamma_k)$ is strictly decreasing (see
Lemma~\ref{lem:tilt}), at most one of $\mathcal{C}_+$,
$\mathcal{C}_-$ will be active for a given row $k$.
Corollary~\ref{cor:slackness}, shows that if the reference row
already satisfies the prescribed bound, the KL
projection onto that halfspace leaves the row unchanged.
If the constraint is active, the scalar $\gamma_k$ is updated by Newton's method (Lemma~\ref{lem:tilt}).
Crucially, relaxing the marginal constraints via $\lambda$ means that
the Sinkhorn scalings $(u,v)$ must be re-solved jointly with the tilt
multipliers $\{\gamma_i\}$: the two sub-problems are coupled because
the tilt changes the effective kernel seen by the Sinkhorn step, and
the Sinkhorn step changes the row masses against which the tilt is
computed.  The complete alternating Bregman procedure is given as
Algorithm~\ref{alg:gap} in Appendix~A.2.


\section{Projection Gap and Diagnostic}
\label{sec:gap}

Let $(X_t)_{t=0}^{T}$ be a real-valued process observed on a
discrete grid $t_k = k\Delta t$, with associated natural filtration $\mathcal{F}_k \coloneqq \sigma(X_0,\ldots,X_k)$.
We interpret the observed path as a discretisation of an underlying continuous-time process, where $N = T/\Delta t$ governs the resolution of the empirical dynamics.

\paragraph{Empirical local transitions.}
To probe local one-step dynamics we construct empirical transition
measures using a rolling window of size $h$.
For each $k \geq h$, let
\begin{equation*}
  \hat{\mu}_k \coloneqq \frac{1}{h}\sum_{s=k-h+1}^{k}\delta_{Z_s},
  \qquad
  \hat{\nu}_k \coloneqq \frac{1}{h}\sum_{s=k-h+1}^{k}\delta_{Y_s},
\end{equation*}
where $Z_s \in \mathcal{F}_{s-1}$ is the chosen predictor and
$Y_s = X_s - X_{s-1}$ is the one-step increment.
These empirical measures provide a finite-sample proxy for the
joint law of predictor states and subsequent increments over the
local window.

\subsection{Marginal relaxation and the projection gap}
\label{sec:gap:relax}

The drift-admissibility constraint set $\mathcal{C}_k \coloneqq \mathcal C_- \cap \mathcal C_+$ may be
\emph{infeasible} for the hard-marginal problem \eqref{eq:hard} even when the underlying process is genuinely admissible.
This occurs for two structurally distinct reasons:

\textbf{(i) Discretisation-induced aggregation.}
    At coarse predictor resolutions, each bin $\mathcal{B}_k$
    pools states $z \in \mathcal{B}_k$ whose true conditional
    means $\mathbb{E}[Y \mid Z = z]$ need not be identical.
    The empirical row mean of the coupling for bin $k$ is
    therefore the bin-averaged conditional mean
    \begin{equation}
      \hat{m}_k
      \;=\;
      \frac{1}{\hat{\mu}_k(\mathcal{B}_k)}
      \int_{\mathcal{B}_k}
      \mathbb{E}[Y \mid Z = z]\,\hat{\mu}_k(dz).
      \label{eq:bin-mean}
    \end{equation}
    For a \emph{memoryless} process, $\mathbb{E}[Y \mid Z = z]
    \approx 0$ for all $z$, so $\hat{m}_k \approx 0$
    irrespective of bin width and the drift constraint remains
    inactive.
    For a process \emph{with memory}, however,
    $\mathbb{E}[Y \mid Z = z]$ varies systematically with $z$:
    states within the same bin share a directional bias, and
    the bin average~\eqref{eq:bin-mean} inherits and preserves
    this bias rather than cancelling it.
    Coarse binning therefore acts as a \emph{low-pass filter on
    the conditional mean structure}: it retains the memory
    signal while smoothing high-frequency noise.
    The only genuine loss occurs when a bin boundary splits a
    region where the conditional mean changes sign, in which
    case the averaging is conservative---it suppresses
    $G_{\lambda,k}$ rather than inflating it, so the diagnostic
    never falsely detects memory that is not present.

\textbf{(ii) Structural incompatibility.}
        The process genuinely violates the prescribed admissible
        class: the conditional mean of the increment distribution
        lies outside the drift box $[d^n_-, d^n_+]$ for a positive
        fraction of predictor states, independently of the
        discretisation resolution.

\begin{remark}[Discretisation as a memory filter]
\label{rem:disc-memory}
Cause~(i) is not a weakness of the diagnostic but a structural
feature: it means that coarser predictor bins \emph{amplify} the
separation between process families with different memory
structures.
For BSS processes, the power kernel introduces long-range
dependence that creates stronger directional persistence within
each predictor bin than fBM at the same equivalent Hurst index.
Coarse binning therefore makes the BSS--fBM separation
\emph{more} visible, not less---a property confirmed empirically
in Section~\ref{sec:exp:hurst}.
\end{remark}

Replacing the hard marginal constraints with soft penalties handles both cases uniformly: the solver is
always well-posed, and the resulting diagnostic reflects the structural distance to admissibility. As such, we adopt the following structure for our diagnostic.


\paragraph{Relaxed-marginal objective.}
Given a cost matrix $C_k$ on the local predictor--increment grid,
define the Gibbs kernel $K_{0,k} \coloneqq \exp(-C_k/\varepsilon)$.
In all experiments we use the quadratic cost
$C_{k,ij} \coloneqq (Y_j - Z_i)^2$,
a smoothness prior that favours cheap local transitions.
The \emph{marginal-relaxed objective} is
\begin{equation}
  \mathcal J_{\varepsilon,\lambda}(T;\,\hat{\mu}_k,\,\hat{\nu}_k)
  \;\coloneqq\;
  \varepsilon \infdiv{T}{K_{0,k}}
  \;+\;
  \lambda\!\left[
    \infdiv{T\mathbf{1}}{\hat{\mu}_k}
    +
    \infdiv{T^{\top}\mathbf{1}}{\hat{\nu}_k}
  \right],
  \label{eq:cJ}
\end{equation}
where $\lambda > 0$ controls how strictly the empirical marginals
are enforced.  The \emph{reference coupling} and
\emph{drift-constrained projection} are then
\begin{equation}
  \Tref \;\coloneqq\;
  \operatorname*{arg\,min}_{T \geq 0}\;
  \mathcal J_{\varepsilon,\lambda}(T;\,\hat{\mu}_k,\,\hat{\nu}_k),
  \qquad
  \Tcon \;\coloneqq\;
  \operatorname*{arg\,min}_{T \in \mathcal{C}_k}\;
  \mathcal J_{\varepsilon,\lambda}(T;\,\hat{\mu}_k,\,\hat{\nu}_k),
  \label{eq:Tref-Tcon}
\end{equation}
where $\mathcal{C}_k$ is the local predictable-box constraint set
(Definition~\ref{def:compat}).
By Lemma~\ref{lem:relaxed-sinkhorn}, $\Tref$ has the scaling form
$\mathrm{diag}(u)\,K_{0,k}\,\mathrm{diag}(v)$.

The complete rolling-window procedure, including the row-wise
tilt step and warm-starting scheme, is given in
Algorithm~\ref{alg:gap}.

\begin{lemma}[Relaxed Sinkhorn as KL projection with soft marginals
{\cite{chizat2018scaling,langbridge2026sinkhorn}}]
\label{lem:relaxed-sinkhorn}
Let $K_0 \in \mathbb{R}_+^{m \times \ell}$ and $\mu, \nu$ be the
observed marginals.
For $\lambda > 0$, the relaxed objective~\eqref{eq:cJ} for $T \in
\mathbb{R}_+$ admits a unique
minimiser of the form $T^* = \mathrm{diag}(u) K_0 \mathrm{diag}(v)$,
for some scaling vectors $(u, v) \in \mathbb{R}_+^m \times
\mathbb{R}_+^\ell$.
\end{lemma}

\begin{lemma}[Row-wise KL projection and exponential tilt]
\label{lem:tilt}
Fix $(u, v)$ and define $\tilde{T}_{kj} := u_k K_{kj} v_j$.
Consider the row-wise KL projection
\begin{equation*}
  \min_{T_{k\cdot} \geq 0} \sum_{j=1}^\ell T_{kj} \log
  \frac{T_{kj}}{\tilde{T}_{kj}} \quad \text{s.t.} \quad
  \sum_{j=1}^\ell g^n(z_k, y_j) T_{kj} = c_k.
\end{equation*}
Assume $g^n(z_k, y_j)$ is finite and non-constant on the support of
$\tilde{T}_{k\cdot}$.
Then the following hold.
\begin{enumerate}[label=(\roman*)]
  \item \textbf{Exponential tilt.} The unique solution has the form
        $T_{kj}(\gamma_k) \propto \tilde{T}_{kj} \exp(-\gamma_k
        g^n(z_k, y_j)/\varepsilon)$, where $\gamma_k$ is the unique
        scalar satisfying $\psi_k(\gamma_k) = c_k$ and $\psi_k(\gamma)
        := \sum_j g^n(z_k, y_j) T_{kj}(\gamma)$.
  \item \textbf{Strong monotonicity.} The map $\psi_k$ is
        $C^\infty$, strictly decreasing, and satisfies
        $\psi_k'(\gamma) = -\varepsilon^{-1} m_k
        \mathrm{Var}_{\pi_k(\gamma)}(g^n(z_k, Y))$.
  \item \textbf{Stability.} The inverse map $\gamma_k =
        \psi_k^{-1}(c_k)$ is Lipschitz: $|\gamma_k(c) -
        \gamma_k(c')| \leq \varepsilon (m_k \sigma_k^2)^{-1} |c -
        c'|$.
  \item \textbf{Numerical root-finding.} The scalar equation
        $\psi_k(\gamma_k) = c_k$ can be solved globally by bisection
        and locally by Newton's method with quadratic convergence.
\end{enumerate}
\end{lemma}

\begin{theorem}[Convergence of relaxed predictable-drift projection]
\label{thm:convergence}
Assume $\varepsilon > 0$, $\lambda > 0$, and that $\mathcal{C}_+$
and $\mathcal{C}_-$ are closed convex sets with nonempty
intersection.
Then the relaxed objective~\eqref{eq:cJ} admits a unique
minimiser for $\Tref$ and $\Tcon$.
Moreover, the iterative scheme combining relaxed Sinkhorn updates
with Bregman projections onto $\mathcal{C}_+$ and $\mathcal{C}_-$
converges to this minimiser.

If additionally $\mathrm{Var}_{\pi_k(\gamma)}(g^n(z_k, Y)) \geq
\sigma_k^2 > 0$ uniformly, then the algorithm converges linearly.
\end{theorem}

\begin{proof}
    The proof of Theorem ~\ref{thm:convergence} is deferred to Appendix \ref{app:proofs}.
\end{proof}

\begin{definition}[Projection gap]
\label{def:gap}
The \emph{projection gap} at time $t$ is
\begin{equation}
  G_{\lambda,k} := \infdiv{T^{\mathrm{proj}}_k}{T^{\mathrm{ref}}_k}.
  \label{eq:gap}
\end{equation}
By strict convexity and attainment, $G_{\lambda,k} = 0$ if and only if
$\Tref \in \mathcal{C}_k$, i.e.\ the unconstrained reference coupling already satisfies theprescribed drift constraints over the window (\textbf{Mode~I}). The non-trivial cases are:

\textbf{Mode II} (Tilt channel, $G_{\lambda,k} > 0$, feasible).
        The set $\mathcal{C}_\mu \cap \mathcal{C}_\nu \cap \mathcal{C}_k$ is non-empty but
        $\Tref \notin \mathcal{C}_k$.  The projection
        $\Tcon$ is obtained by exponential tilting of the
        reference rows (Lemma~\ref{lem:row-tilt}): each active row
        is reweighted to shift its conditional mean to the nearest
        admissible boundary, preserving the row support.
        This mode is characteristic of processes whose increment
        distribution is \emph{feasibly} admissible but whose
        entropic optimum lies outside the drift box.

\textbf{Mode III} (Mass reallocation, $G_{\lambda,k} > 0$, infeasible).
        The hard-marginal constraint set
        $\mathcal{C}_\mu \cap \mathcal{C}_\nu \cap \mathcal{C}_k$
        is empty for some rows: no coupling with the prescribed
        marginals can satisfy the drift box for those predictor
        states. Under the soft-marginal objective~\eqref{eq:cJ},
        the solver responds by reducing the mass assigned to
        infeasible predictor states, effectively down-weighting
        windows where the empirical increment distribution is
        structurally incompatible with the admissible class.
        This mode is characteristic of rough processes
        ($H < 0.5$) at fine mesh resolutions, where the
        increment variance dominates the drift envelope.
\end{definition}

By Sanov's theorem,
$G_{\lambda,k}$ is the exponential rate at which the observed law is improbable under the nearest admissible law. Larger values of $G_{\lambda,k}$ therefore represent stronger evidence of structural incompatibility.



\begin{lemma}[Quadratic lower bound on row projection gap]
\label{lem:quadratic-lb}
Fix a row $k$ with row mass $m_k \geq 0$, and let $T_k(\gamma)$ be
the exponential tilt of a reference row $\tilde{T}_k$ as in
Lemma~\ref{lem:tilt}.
Then
\begin{equation*}
  \frac{d}{d\gamma} \infdiv{T_k(\gamma)}{\tilde{T}_k} =
  \frac{m_k \gamma}{\varepsilon^2}
  \mathrm{Var}_{\pi_k(\gamma)}(g^n(z_k, Y)).
\end{equation*}
Consequently, if $\mathrm{Var}_{\pi_k(\gamma)}(g^n(z_k, Y)) \geq
\sigma_k^2 > 0$ for $0 \leq \gamma \leq \bar{\gamma}$, then
\(
  G_{\lambda,k} \;\geq\; \frac{m_k \sigma_k^2}{2\varepsilon^2}
  \gamma^2.
\)
\end{lemma}

\begin{proposition}[Monotonicity of the projection gap]
\label{prop:monotone}
Fix \(k\) and \(\varepsilon,\lambda>0\), and let \(\{\mathcal C_k^{(r)}\}_{r\geq 1}\) be a nested family of local predictable-box constraint sets satisfying
$
\mathcal C_k^{(r+1)} \subseteq \mathcal C_k^{(r)}.
$

For each \(r\), define
$
{\Tcon}^{(r)}
\coloneqq
\operatorname*{arg\,min}_{T\in\mathcal C_k^{(r)}}
\mathcal J_{\lambda}(T;\hat{\mu}_k,\hat{\nu}_k),
$
and let
$
G_{\lambda,k}^{(r)}
\coloneqq
\infdiv{{\Tcon}^{(r)}}{\Tref}.
$

Suppose that, at each constrained optimum, complementary slackness and monotonicity of the drift constraints imply that the active constraint set contains at most one affine constraint (Lemma \ref{lem:tilt}).
Then, by the KL Pythagorean identity,
$$
\mathcal C_k^{(r+1)} \subseteq \mathcal C_k^{(r)} \Rightarrow
\mathcal J_{\varepsilon, \lambda}({\Tcon}^{(r+1)})
\geq
\mathcal J_{\varepsilon, \lambda}({\Tcon}^{(r)}) \Rightarrow
G_{\lambda,k}^{(r+1)}
\geq
G_{\lambda,k}^{(r)}.
$$
Thus, tightening the predictable-box constraint cannot decrease the KL divergence of the optimal coupling from the unconstrained reference coupling.
\end{proposition}


\subsection{Calibration against a Brownian null}
\label{sec:calibration}
To calibrate the drift envelope in a resolution-invariant manner,
we use the reflection principle for standard Brownian motion.
Let $M_W \coloneqq \sup_{0 \leq s \leq W} B_s$ denote the running
supremum of a standard Brownian motion over a window of duration
$h\Delta t$.  Then
\begin{equation}
  \mathbb{P}(M_W \geq y)
  \;=\;
  2\,\mathbb{P}(B_W \geq y)
  \;=\;
  2\!\left(1 - \Phi\!\left(\frac{y}{\sqrt{h\Delta t}}\right)\right),
  \label{eq:reflection}
\end{equation}
where $\Phi$ denotes the standard normal CDF.
Setting $y = c\sqrt{h\Delta t}$ gives a drift envelope
\begin{equation}
  d_{\pm}(x_i) \;\equiv\; \pm\,c\sqrt{h \cdot \Delta t},
  \label{eq:envelope}
\end{equation}
where the critical coefficient $c = \Phi^{-1}(1 - \alpha/2)$ is
invariant to both window length and sampling rate for a targeted
Brownian violation probability $\alpha$.
Throughout this paper we adopt the canonical calibration
$c = 0.50$, corresponding to $\alpha = 2(1-\Phi(0.5)) \approx 0.617$:
approximately $62\%$ of Brownian windows activate the constraint
boundary, ensuring separation between process classes while
remaining a deliberately loose threshold.

\section{Experiments}
\label{sec:experiments}


We evaluate the diagnostic on synthetic stochastic processes
spanning a range of local regularity, memory structure, and
jump behaviour\footnote{Code to reproduce all experiments is available at \url{https://anonymous.4open.science/r/drift-admissibility-F36B/}}.
The primary comparison uses fractional Brownian motion (fBM) and
Brownian semistationary (BSS) processes with a power convolution
kernel, both parametrised by an equivalent Hurst index
$H \in [0.10, 0.90]$.

Table~\ref{tab:processes} summarises the processes, and unless otherwise stated, all experiments use $N=400$ observations
($\Delta t = 0.0025$), window size $h=30$ ($h\Delta t = 0.075$),
entropic regularisation $\varepsilon = 20/N$, marginal penalty
$\lambda = 20$, envelope scale $c = 0.50$ ($\alpha \approx 62\%$
Brownian violation probability), and $10$ independent MC trials
(error bars: $95\%$ CI).
\begin{table}[ht]
\centering
\caption{%
  Synthetic processes used in the primary experiments.
  All processes are simulated over $t \in [0,1]$ with
  $\Delta t = 1/N$.%
}
\label{tab:processes}
\begin{tabular}{llll}
\toprule
\textbf{Process} & \textbf{Evolution} & \textbf{Parameters} \\
\midrule
Brownian motion
  & $dX_t = \sigma\,dW_t$
  & $\sigma = 1.0$ \\
Fractional BM (fBM)
  & $X_t = B^H_t$
  & $H \in (0,1)$ \\
BSS (power kernel)
  & $\begin{array}{l} X_t = \kappa \int_0^t g(t-s)\,dW_s \\ g(x) = x^{\alpha}(1+x)^{-\gamma-\alpha} \end{array}$
  & $\begin{array}{l} \alpha \in (-0.5, 0.5),\\ \gamma = \kappa = 1.0 \end{array}$ \\
\bottomrule
\end{tabular}
\end{table}

\subsection{Mean projection gap across the Hurst spectrum}
\label{sec:exp:hurst}

\begin{figure}[t]
  \centering
  \includegraphics[width=0.45\linewidth]{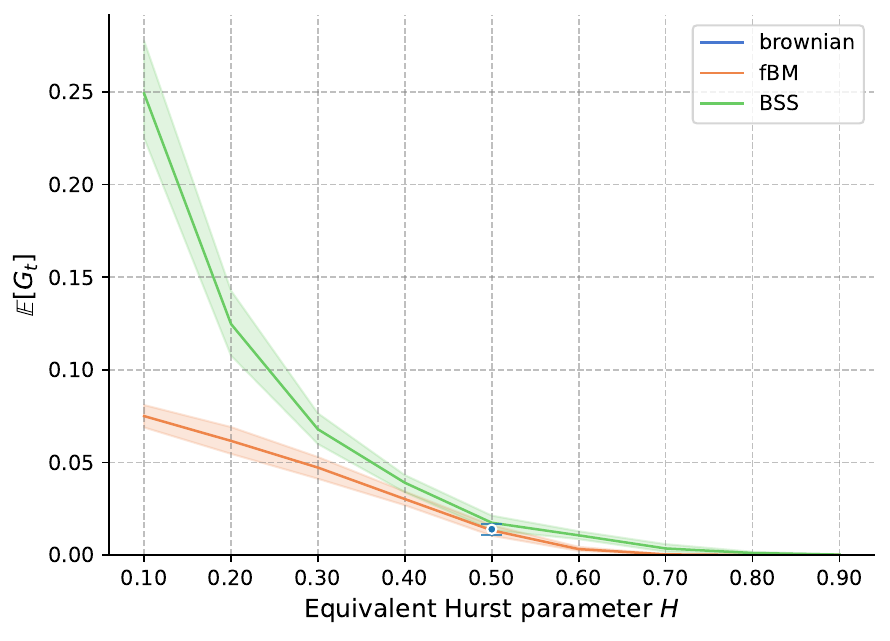}
  \caption{%
    Mean projection gap $\mathbb{E}[G_{\lambda,t}]$ against
    equivalent Hurst parameter $H$ for Brownian motion, fBM, and
    BSS (power kernel).
    Shaded ribbons: pointwise 95\% CI over 10 MC trials.%
  }
  \label{fig:hurst-sweep}
\end{figure}

Figure~\ref{fig:hurst-sweep} reports the expected projection gap
$\mathbb{E}[G_{\lambda,t}]$ against the equivalent Hurst index for
fBM and BSS processes, together with the Brownian reference.
Three regimes emerge cleanly, each corresponding to a distinct
mode of the projection gap characterised in
Section~\ref{sec:gap:relax}.

\textbf{Rough regime ($H < 0.5$).}
The increment variance is proportional to $(\Delta t)^{2H}$ with
$2H < 1$.
For $\Delta t \ll 1$, this exceeds the Brownian null variance
$\Delta t$, causing systematic constraint violations and a
monotonically increasing gap as $H$ decreases due to joint contributions of \emph{Mode~II} (tilt) and \emph{Mode~III} (mass reallocation).

\textbf{Brownian reference ($H = 0.5$).}
The Brownian process yields a small but consistent gap across all
ten MC trials, overlaying the fBM curve at $H=0.5$.
These violations are an intentional consequence of the $62\%$
calibration (Section~\ref{sec:calibration}) and ensure
separation from smooth processes. This corresponds to
\emph{Mode~II} (tilt): the empirical increment
distribution is feasibly admissible, but the entropic optimum
lies just outside the drift box, requiring a small exponential
tilt to enforce the constraint.

\textbf{Smooth regime ($H > 0.5$).}
The increment variance is $\propto (\Delta t)^{2H}$ with $2H > 1$,
so increments are smaller than the Brownian null for $\Delta t \ll 1$.
The drift envelope is non-binding and the gap collapses to \emph{Mode~I} (inactive constraint) as $H \to 1$: the
unconstrained reference coupling already satisfies the prescribed
drift class and no deformation is required.

\textbf{BSS--fBM separation.}
At rough indices ($H < 0.5$), BSS processes exhibit structurally
higher projection gaps than fBM counterparts at the same equivalent
$H$.
Since fBM and BSS share the same local scaling exponent, no
diagnostic that depends on the process only through its local
regularity---such as quadratic variation or an estimated Hurst
exponent---can distinguish them.
The separation is attributable to the kernel-dependent memory
structure of BSS, which creates stronger directional persistence
within each predictor bin (Remark~\ref{rem:disc-memory}).
Note also that the BSS curve at $H=0.5$ remains above the Brownian
reference, providing empirical evidence that a nominal scaling
exponent of $H=0.5$ is insufficient to guarantee martingale
admissibility for non-Gaussian processes with long-range dependence.

\subsection{Sensitivity and grid refinement}
\label{sec:exp:ablation}

\begin{figure}[t]
  \centering
  \includegraphics[width=\linewidth]{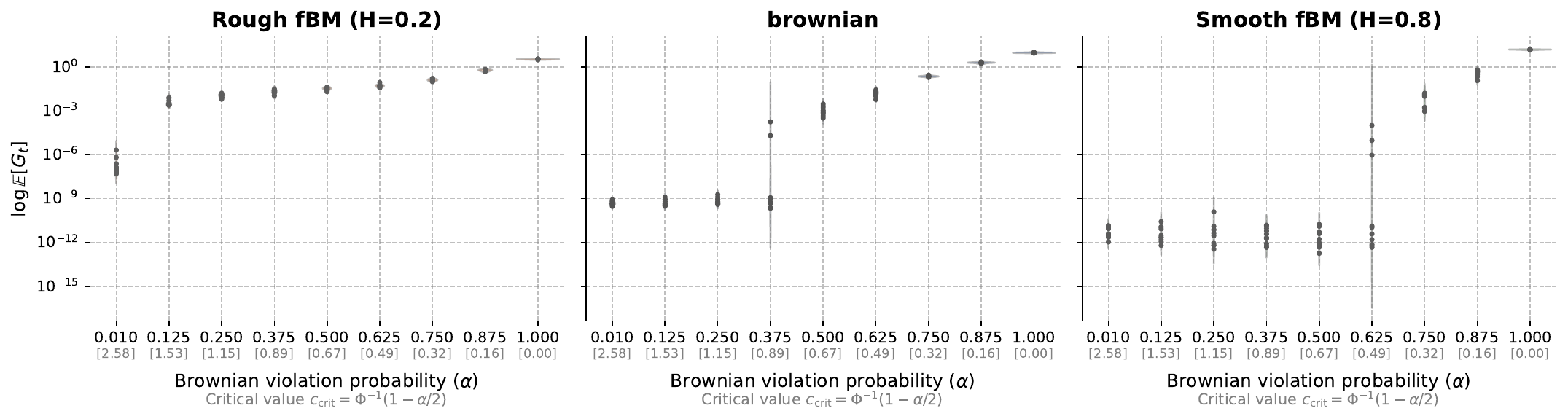}
  \caption{%
    Natural logarithm of the mean gap
    $\log\mathbb{E}[G_{\lambda,t}]$ against the Brownian violation
    probability $\alpha$ for rough fBM ($H=0.2$), Brownian motion,
    and smooth fBM ($H=0.8$).
    Three panels, one per process; secondary axis shows the
    corresponding critical value
    $c_{\mathrm{crit}} = \Phi^{-1}(1-\alpha/2)$.%
  }
  \label{fig:ablation-c}
\end{figure}

\textbf{Sensitivity to drift envelope scale.} Figure~\ref{fig:ablation-c} reports $\log\mathbb{E}[G_{\lambda,t}]$
against the Brownian violation probability $\alpha$ for rough fBM
($H=0.2$), Brownian motion, and smooth fBM ($H=0.8$).
The profile for each process is non-increasing in $c$ (equivalently,
non-decreasing in $\alpha$), consistent with
Proposition~\ref{prop:monotone}.

The mean projection gap demonstrates distinct bimodality for all
three process types, directly reflecting the three-mode
characterisation of Section~\ref{sec:gap:relax}.
For $\alpha \leq 0.25$, the drift envelope is wide enough that
constraints are inactive for Brownian motion (\textbf{Mode~I}):
the reference coupling already lies in $\mathcal{C}_k$ and
$G_{\lambda,k} \approx 0$.
For $\alpha \in [0.25, 0.625]$, the modality shift reflects the
onset of \textbf{Mode~II} (tilt channel) violations: the
constraint becomes active for an increasing fraction of windows,
and the solver responds with exponential tilting of the reference
rows.
For $\alpha > 0.625$, the gap increases exponentially as the
envelope narrows further, consistent with the exponential-tilt
form of the Bregman projection established in
Theorem~\ref{thm:convergence}.
The rough fBM ($H = 0.2$) enters Mode~II at a much lower $\alpha$
than Brownian motion, since its large increment variance
systematically exceeds the drift envelope even at loose
calibrations.
The smooth fBM ($H = 0.8$) requires the tightest envelope before
any violations appear, consistent with its sub-Brownian increment
variance.

\begin{figure}[t]
  \centering
  \includegraphics[width=\linewidth]{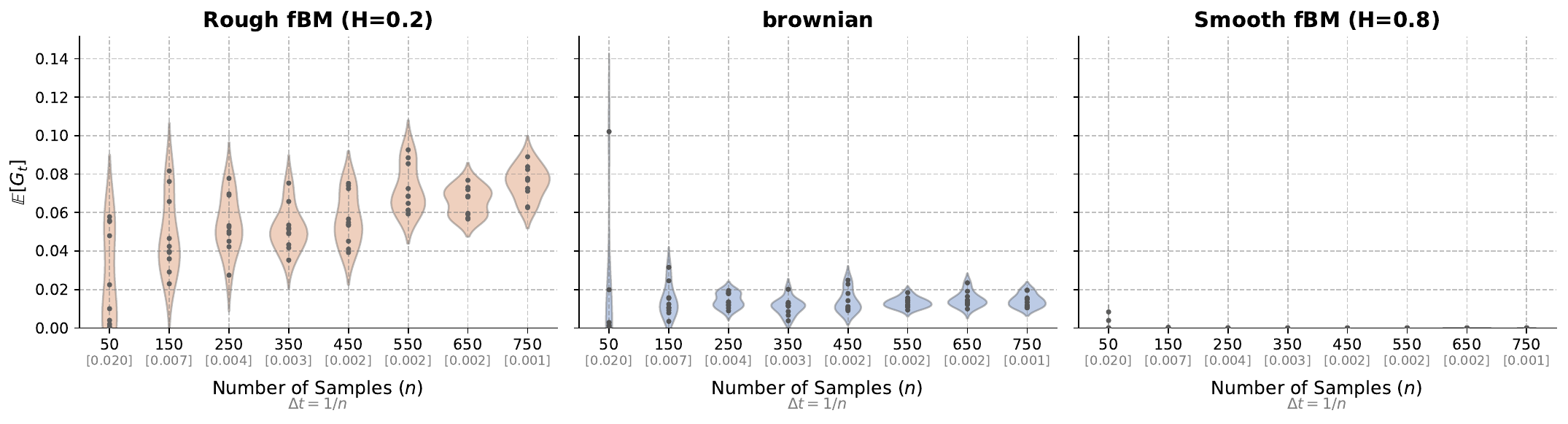}
  \caption{%
    Mean gap $\mathbb{E}[G_{\lambda,t}]$ against grid
    resolution $n$ ($\Delta t = 1/n$) for rough fBM ($H=0.2$),
    Brownian motion, and smooth fBM ($H=0.8$).
    Three panels, one per process; secondary axis shows $\Delta t$.
  }
  \label{fig:refinement}
\end{figure}

\textbf{Behaviour under grid refinement.} Figure~\ref{fig:refinement} reports $\mathbb{E}[G_{\lambda,t}]$
against the number of samples $n \in [50, 750]$ (equivalently
$\Delta t \in [0.02, 0.001]$) for rough fBM, Brownian motion, and
smooth fBM, with $c = 0.5$ and $h = 30$ fixed.

The three regimes show qualitatively distinct scaling behaviour,
again directly traceable to the three-mode taxonomy.
The \textbf{Brownian regime} is stable across resolutions for
$n > 150$: the gap converges to a consistent level reflecting the
deliberate 62\% calibration (Mode~II), while the high variance at
$n = 50$ reflects the large window-to-sample ratio ($h/n = 0.6$)
rather than any structural violation.
The \textbf{rough regime} shows a broadly increasing gap with
resolution: as $\Delta t$ decreases, the path increment grows
$\propto (\Delta t)^{0.4}$, causing more windows to enter
Mode~III (mass reallocation) as the increment variance increasingly
dominates the drift envelope.
This increase is modulated by the rough process's mean-reversion,
which encourages reversion within the observation window and
partially offsets the constraint violations.
The \textbf{smooth regime} collapses to $\mathbb{E}[G_{\lambda,t}]
\approx 0$ for $n > 50$ (Mode~I): the smoothing dominates to
produce sub-Brownian increments for which the drift envelope is
non-binding at all tested resolutions.
At $n = 50$, some non-zero gaps appear because the increment of
the smooth process is only $\approx 10\times$ smaller than the
Brownian null, which can produce occasional Mode~II violations at
very coarse resolutions.

\subsection{Solver validation}
\label{sec:exp:solver}


We validate the Bregman--Dykstra implementation against an
equivalent formulation solved with \textsc{cvxpy} (MOSEK backend)
over 200 independent transport problems. All problems use the same cost matrix, entropic regularisation, marginal penalties, and predictable-drift constraints. The Bregman plan is KL-minimal from $T^{\mathrm{ref}}$ by
construction, so all reported gaps are lower bounds on the
equivalent MOSEK scores. Table~\ref{tab:solver} reports four complementary diagnostics.

\begin{table}[t]
\centering
\caption{%
  Quantitative comparison of the proposed Bregman--Dykstra
  projection algorithm against \textsc{cvxpy} (MOSEK) over 200
  independent transport problems.%
}
\label{tab:solver}
\begin{tabular}{lrrrr}
\toprule
\small
\textbf{Metric} & \textbf{Mean} & \textbf{Std} & \textbf{Min} & \textbf{Max} \\
\midrule
Relative objective gap
  $\frac{|\mathcal J_{\varepsilon,\lambda}(T) - \mathcal J_{\varepsilon,\lambda}(T^{\mathrm{MOSEK}})|}{|\mathcal J_{\varepsilon,\lambda}(T^{\mathrm{MOSEK}})|}$
  & 0.0203 & 0.0112 & 0.0016 & 0.0656 \\
Max plan discrepancy
  $\max_{ij}\left|\frac{T_{ij}}{|T|_1}
  - \frac{T^{\mathrm{MOSEK}}_{ij}}{|T^{\mathrm{MOSEK}}|_1}\right|$
  & 0.0022 & 0.0016 & 0.0003 & 0.0106 \\
Max constraint violation
  & \small{$2.17 \times 10^{-5}$} & \small{$2.76 \times 10^{-4}$} & $0$ & $0.0035$ \\
Mean absolute drift error
  $\frac{1}{n}\sum_i |d_i - d^{\mathrm{MOSEK}}_i|$
  & 0.0033 & 0.0020 & 0.0003 & 0.0099 \\
\bottomrule
\end{tabular}
\end{table}

Although the two methods occasionally converge to slightly different
optimal transport plans (reflecting the non-uniqueness of the
primal solution when multiple couplings achieve the same objective),
the resulting conditional drift functions are nearly
indistinguishable and satisfy the prescribed admissibility
constraints to high accuracy.
The remaining differences reflect the selection of numerically
equivalent optimal couplings rather than a loss of optimisation
accuracy.
Runtime scaling curves and Dolan--Moré performance profiles are
reported in Appendix~\ref{app:complexity}.


\section{Discussion and Limitations}
\label{sec:discussion}


Our experiments demonstrate that the projection gap captures
differences between stochastic regimes that are not explained by
local regularity alone.
In particular, the BSS--fBM separation at identical Hurst indices
(Figure~\ref{fig:hurst-sweep}) provides direct evidence that the
diagnostic is sensitive to kernel-dependent memory structure beyond
second-order statistics.
The mechanistic explanation---that coarse predictor bins act as a
low-pass filter preserving directional memory signals---suggests
that the diagnostic may be more sensitive to long-range dependence
at coarser resolutions, a property that warrants further
theoretical investigation.

\textbf{Limitations.}
The projection gap depends jointly on the observed data, the admissible
class, the filtration discretisation, and the transport geometry---a
feature, not a weakness, since admissibility is inherently relative to a
modelling assumption, but one that means scores from different cost
matrices, envelope scales, or window sizes are not directly comparable.
The computational complexity of the diagnostic is
$\mathcal{O}((N-h) \cdot h^3 \varepsilon^{-1} \log \varepsilon^{-1})$,
which is manageable for the window sizes and resolutions considered
here but may require subsampling or parallelisation for very long
time series.

\textbf{Future work.}
The immediate extension of the present work is to characterise the
null distribution of $G_{\lambda,t}$ under a specified null model,
enabling a calibrated hypothesis test.
Such a test would require a joint asymptotic analysis of the
empirical transition discretisation, the rolling-window
approximation, the entropic regularisation, and the drift-envelope
scaling.
For moment-inequality null classes of the form considered here, the
limiting distribution is expected to be of chi-bar-squared type
\citep{andrews2013inference, chetverikov2018adaptive}, and
least-favourable calibration would be required to control size
uniformly.
A second direction is to learn the admissible class from data by
parametrising the drift bounds and alternating between estimation
and projection, extending the method toward data-driven structure
discovery at the cost of introducing nonconvexity and
identifiability challenges; multivariate extensions with richer constraint classes (factor structure, cross-variable dependencies) would further broaden
applicability.

\section{Conclusion}
In this work, we have presented a  structure-aware diagnostic for assessing whether empirical time-series dynamics are compatible with a prescribed predictable drift class. Our approach casts this as a finite-dimensional projection problem over empirical transition laws, which can be solved efficiently using iterative Bregman projections. The resulting projection gap $G_{\lambda,t}$, calibrated against a Brownian reference null, captures systematic deviations from the predictable drift class and separates process classes that are indistinguishable by second-order statistics alone.


\bibliographystyle{abbrvnat}
\bibliography{bibl}

\newpage
\appendix
\renewcommand{\thefigure}{A\arabic{figure}} 
\setcounter{figure}{0} 
\renewcommand{\thelemma}{A\arabic{lemma}} 
\renewcommand{\thecorollary}{A\arabic{corollary}} 
\setcounter{lemma}{0} 

\section{Stochastic Analysis Background}\label{app:stochastic-analysis}

A {\it stochastic basis} is a quadruple
$(\Omega,\mathcal F,\mathbb F,\mathbb P)$,
where $(\Omega,\mathcal F,\mathbb P)$ is a probability space and
where $\mathbb F:=(\mathcal F_t)_{t\in[0,T]}$ denotes a filtration, that is, an increasing family of sub-\(\sigma\)-fields
of \(\mathcal F\):
\[
\mathcal F_s\subseteq \mathcal F_t\subseteq \mathcal F,
\qquad 0\le s\le t\le T.
\]
Throughout the paper we take the filtration to be right-continuous, meaning that
\[
\mathcal F_t=\bigcap_{u>t}\mathcal F_u,
\qquad 0\le t<T.
\]
The \(\sigma\)-field \(\mathcal F_t\) represents the information available up to time
\(t\). Accordingly, martingale, local-martingale, predictability, and semimartingale
properties are always understood relative to the chosen stochastic basis.

A process $X=(X_t)_{t\in[0,T]}$ on this basis is
said to be $\mathbb F$-{\it adapted} if $X_t$ is $\mathcal F_t$-measurable for every
$t\in [0,T]$ (often written $X_t\in \mathcal F_t$). A process is $\mathbb F$-{\it predictable} if it is measurable with respect to the
predictable sigma-field $\mathcal P$ on $\Omega\times [0,T]$, generated by the
left-continuous adapted processes; in particular, left-continuous
adapted processes are predictable. A {\it càdlàg}, or RCLL, process is a process whose sample paths are right-continuous
and admit left limits.
A process $X$ is an $\mathbb F$-{\it martingale} if it is an adapted integrable process such that
\[
\mathbb E[X_s \mid \mathcal F_t]=X_t, \qquad 0\leq t\leq s\leq T.
\]
Equivalently, conditional on the information available at time $t$, the future
value of the process has expectation equal to its present value. A {\it local martingale} on \([0,T]\) is an adapted process \(M\) for which there exists
a sequence of stopping times \((\tau_k)_k\) with \(\tau_k\uparrow T\) a.s. such that \(M^{\tau_k}\) is a martingale on \([0,T]\) for every \(k\).
A real-valued, càdlàg $\mathbb F$-adapted process $X$
is a {\it semimartingale}, with respect to \(\mathbb F\), if it admits a decomposition 

\[
X_t=X_0+M_t+A_t,\qquad t\in[0,T],
\]
where \(M\) is an \(\mathbb F\)-local martingale and \(A\) is an \(\mathbb F\)-adapted
process of finite variation. 
In this paper, we focus on the {\it special semimartingale} case, in which such a decomposition can be chosen with \(A\) predictable. See \cite{JacodShiryaev} for background and for the refinement of this structure in terms of predictable characteristics for Itô
semimartingales. For the purposes of the present paper, the essential point is the
special semimartingale separation between conditionally centered local-martingale
innovations and a predictable finite-variation component.

\section{Algorithm Pseudocode}
\label{app:alg}
Algorithm~\ref{alg:gap} gives the complete rolling-window
procedure for computing the projection gap sequence
$\{G_{\lambda,k}\}_{k=h}^{N-1}$ from a single observed path.
The three structural cases correspond directly to the three-mode
characterisation of Definition~2: rows satisfying
$\psi_k(0) \in [d_-, d_+]$ contribute $G_{\lambda,k} = 0$
(Mode~I); rows requiring a tilt ($\gamma_i^\star \neq 0$) but
with a non-empty feasible set contribute via the exponential-tilt
channel (Mode~II); and rows for which no coupling with the
prescribed marginals can satisfy the drift box are handled by the
soft-marginal objective, which reduces their mass contribution
(Mode~III).

\begin{algorithm}[H]
\caption{Projection Gap Diagnostic}
\label{alg:gap}
\begin{algorithmic}[1]
\small

\Require Observed path $X_{t_0}, \ldots, X_{t_N}$;
         window size $h$;
         envelope scale $c$;
         entropic regularisation $\varepsilon > 0$;
         marginal penalty $\lambda > 0$;
         convergence tolerance $\delta > 0$
\Ensure  Projection gap sequence
         $\bigl\{G_{\lambda,k}\bigr\}_{k=h}^{N-1}$

\medskip

\State Set drift envelope half-width based on Brownian calibration: \Comment{Section~\ref{sec:calibration}}
\[
  d_{\pm}(x_i) \;\equiv\; \pm\, c\sqrt{h \cdot \Delta t}
\] 

\medskip

\For{$k = h$ \textbf{to} $N-1$}

  \medskip

  \State Build rolling-window empirical measures over $[k-h+1,\, k]$: \Comment{Section~\ref{sec:gap:relax}}
  \[
    \hat{\mu}_k \;:=\; \frac{1}{h}\sum_{s=k-h+1}^{k} \delta_{Z_s},
    \qquad
    \hat{\nu}_k  \;:=\; \frac{1}{h}\sum_{s=k-h+1}^{k} \delta_{Y_s}
  \]
  \quad\:\: where $Z_s \in \mathcal{F}_{s-1}$ is the predictor and
  $Y_s = X_s - X_{s-1}$ is the one-step increment.

  \medskip

  \State Discretise predictor grid $\{z_i\}_{i=1}^{m}$ and
         increment grid $\{y_j\}_{j=1}^{\ell}$ from the window data.

  \State Set quadratic cost matrix:
  \(
    C_{k,ij} \;\equiv\; (Y_j - Z_i)^2
  \)

  \State Set Gibbs kernel:
  \(
    K_{0,k} \;:=\; \exp(-C_k / \varepsilon)
  \)

  \medskip

  \State Compute $\Tref$ by relaxed Sinkhorn iterations until
         convergence: \Comment{Lemma~\ref{lem:relaxed-sinkhorn}}
  \[
    \Tref \;:=\;
    \argmin_{T \geq 0}\;
    \varepsilon\,\infdiv{T}{K_{0,k}}
    + \lambda\!\left[
        \infdiv{T\mathbf{1}}{\hat{\mu}_k}
        + \infdiv{T^{\!\top}\mathbf{1}}{\hat{\nu}_k}
      \right]
  \]
  \quad\:\: Retain the converged Sinkhorn scalings
  $(u^{\star}, v^{\star}) \in \mathbb{R}^m_+ \times \mathbb{R}^\ell_+$.

  \medskip

  \State Initialise Sinkhorn scalings:
         $u \leftarrow u^{\star}$,\; $v \leftarrow v^{\star}$
         \Comment{warm-start from $\Tref$}

  \State Initialise tilt multipliers:
         $\gamma_i \leftarrow 0$ for all $i = 1, \ldots, m$

  \Repeat \Comment{outer Bregman loop}

    \medskip

    \State Form the tilt-adjusted kernel:
    \[
      \widetilde{K}_{ij}
      \;:=\;
      K_{0,k,ij}
      \exp\bigl(-\gamma_i\,g^n(z_i, y_j)/\varepsilon\bigr),
      \quad i=1,\ldots,m,\; j=1,\ldots,\ell
    \]

    \State Soft-marginal Sinkhorn updates on $\widetilde{K}$:
    \Comment{Lemma~\ref{lem:relaxed-sinkhorn}}
    \[
      u \;\leftarrow\;
        \left(\frac{\hat{\mu}_k}{\widetilde{K}\,v}\right)^{\!\lambda/(\lambda+\varepsilon)},
      \qquad
      v \;\leftarrow\;
        \left(\frac{\hat{\nu}_k}{\widetilde{K}^{\!\top} u}\right)^{\!\lambda/(\lambda+\varepsilon)}
    \]

    \State Set current coupling iterate:
    $\Tcon \leftarrow \mathrm{diag}(u)\,\widetilde{K}\,\mathrm{diag}(v)$

    \medskip

  \For{each predictor bin $i = 1, \ldots, m$}

    \State Compute current row moment:
    \[
      \psi_k(\gamma_i) \;=\; \sum_{j=1}^{\ell}
        g^n(z_i, y_j)\,[\Tcon]_{ij}
    \quad \text{where} \quad  g^n(z,y) = y_j - z_i
    \]

    \If{$\psi_k(\gamma_i) \notin \bigl[d_-(z_i),\; d_+(z_i)\bigr]$}
      \Comment{Constraint active: Corollary~\ref{cor:slackness}}

      \State Set boundary target:
      \(
        d_i \;=\;
        \begin{cases}
          d_-(z_i) & \text{if } \psi_k(0) < d_-(z_i) \\
          d_+(z_i) & \text{if } \psi_k(0) > d_+(z_i)
        \end{cases}
      \)

      \State Update $\gamma_i$ by one Newton step on $\psi_k(\gamma) = d_i$:
      \Comment{Lemma~\ref{lem:tilt},(ii,iv)}
        \[
          \gamma_i \;\leftarrow\;
          \gamma_i
          \;-\;
          \frac{\psi_k(\gamma_i) - c_i}{\psi_k'(\gamma_i)}
        \quad\text{where}\quad \psi_k'(\gamma_i) = -\varepsilon^{-1} m_i
        \mathrm{Var}_{\pi_k(\gamma_i)}\!\bigl(g^n(z_i, Y)\bigr) < 0
        \]


    \Else
      \Comment{Row already admissible, Corollary~\ref{cor:slackness}}
      \State $\gamma_i \leftarrow 0$

    \EndIf

  \EndFor

  \Until{$\max_i |\psi_k(\gamma_i) - d_i^{\mathrm{active}}| \leq \delta$
         \emph{and} Sinkhorn marginal residuals $\leq \delta$}
         
  \Comment{joint convergence; guaranteed by Theorem~\ref{thm:convergence}}

  \medskip
  \State Compute projection gap 
    \(
    G_{\lambda,k} \;:=\; \infdiv{\Tcon}{\Tref}
    \) \Comment{Definition~\ref{def:gap}}

\EndFor

\State \Return $\bigl\{G_{\lambda,k}\bigr\}_{k=h}^{N-1}$

\end{algorithmic}
\end{algorithm}

\paragraph{Complexity.}
Each outer iteration (line~3) requires one relaxed Sinkhorn solve
and $m$ one-dimensional root-finding steps.
The Sinkhorn solve converges in $\mathcal{O}(\varepsilon^{-1}
\log \varepsilon^{-1})$ iterations, each costing $\mathcal{O}(m\ell)$
operations; the root-finding steps each cost $\mathcal{O}(\ell
\log(1/\delta))$ for bisection tolerance $\delta$.
The overall complexity of Algorithm~\ref{alg:gap} is therefore
$\mathcal{O}\bigl((N-h)\cdot h^3\varepsilon^{-1}\log\varepsilon^{-1}\bigr)$,
where the $h^3$ factor reflects the $m = \ell = h$ discretisation
used throughout the experiments.
The outer loop is embarrassingly parallel: windows share no
information and can be evaluated concurrently, providing significant
acceleration when $N \gg h$.
To avoid redundant computation, the constrained solver
$\Tcon$ is warm-started from the Kantorovich potentials
$(u, v)$ recovered during the $\Tref$ solve.

\section{Supplementary Lemmas and Proofs}
\subsection{KL Projection for Conditional Bounds}
\begin{lemma}[Row-wise exponential tilt and scalar dual variable]\label{lem:row-tilt} Fix a row index \(k\), and write $g_{kj}:=g_n(z_k,y_j), j=1,\ldots,\ell.$ 
Let \(\widetilde T_{k\cdot}\in\mathbb R_+^\ell\) be a strictly positive reference row with total mass
$m_k:=\sum_{j=1}^{\ell}\widetilde T_{kj}>0$, and assume that \(g_{kj}\) is finite and nonconstant on the support of
		\(\widetilde T_{k\cdot}\). For \(\gamma\in\mathbb R\), define the tilted row
		\[
		T_{kj}(\gamma)
		:=
		m_k\,
		\frac{
			\widetilde T_{kj}\exp\!\left(-\gamma g_{kj}/\varepsilon\right)
		}{
			\sum_{r=1}^{\ell}\widetilde T_{kr}
			\exp\!\left(-\gamma g_{kr}/\varepsilon\right)
		},
		\qquad j=1,\ldots,\ell.
		\]
		Then \(\sum_jT_{kj}(\gamma)=m_k\), and the row moment $\psi_k(\gamma):=\sum_{j=1}^{\ell}g_{kj}T_{kj}(\gamma)$
		is continuous and strictly decreasing. Moreover,
		$\lim_{\gamma\to+\infty}\psi_k(\gamma)
		=m_k\min_j g_{kj}$ and 
		$\lim_{\gamma\to-\infty}\psi_k(\gamma)
		=m_k\max_j g_{kj}.$
		Consequently, for every
		$
		c_k\in (m_k\min_j g_{kj},\,m_k\max_j g_{kj})
		$
		there exists a unique scalar \(\gamma_k\in\mathbb R\) such that $\psi_k(\gamma_k)=c_k.$ The row \(T_{k\cdot}(\gamma_k)\) is the unique solution of
		\[
		\min_{q\in\mathbb R_+^\ell}
		\sum_{j=1}^{\ell}\left(q_j\log\frac{q_j}{\widetilde T_{kj}}-q_j+
		\widetilde T_{kj}\right)
		\]
		subject to
		$
		\sum_{j=1}^{\ell}q_j=m_k,$ and $
		\sum_{j=1}^{\ell}g_{kj}q_j=c_k.
		$\label{lem:exponential-tilt}
	\end{lemma}

\begin{proof}[Proof of Lemma \ref{lem:exponential-tilt}]
    The normalisation in the definition of \(T_{k\cdot}(\gamma)\) gives
    \(\sum_jT_{kj}(\gamma)=m_k\). Define the probability weights
    \[
    \pi_{kj}(\gamma)
    :=
    \frac{
        \widetilde T_{kj}\exp\!\left(-\gamma g_{kj}/\varepsilon\right)
    }{
        \sum_{r=1}^{\ell}\widetilde T_{kr}
        \exp\!\left(-\gamma g_{kr}/\varepsilon\right)
    }.
    \]
    Then
    \[
    \psi_k(\gamma)=m_k\sum_{j=1}^{\ell}g_{kj}\pi_{kj}(\gamma).
    \]
    Since the support is finite and all \(g_{kj}\) are finite, \(\psi_k\) is smooth. Differentiating gives
    \[
    \psi_k'(\gamma)
    =-\frac{m_k}{\varepsilon}
    \operatorname{Var}_{\pi_k(\gamma)}(g_{kY}).
    \]
    By the nonconstancy assumption, the variance is strictly positive for all finite \(\gamma\),
    so \(\psi_k\) is strictly decreasing and continuous. As \(\gamma\to+\infty\), the
    probability vector \(\pi_k(\gamma)\) concentrates on the minimizers of \(g_{kj}\), and
    as \(\gamma\to-\infty\), it concentrates on the maximizers. This proves the two endpoint
    limits. Existence and uniqueness of \(\gamma_k\) for any interior target \(c_k\) then follow
    from the intermediate value theorem and strict monotonicity.
    
    It remains to identify the minimizer. Introduce Lagrange multipliers \(\alpha\) and
    \(\eta\) for the mass and moment constraints. The first-order condition for an interior
    minimizer is
    \[
    \log\frac{q_j}{\widetilde T_{kj}}+\alpha+\eta g_{kj}=0,
    \]
    so
    \[
    q_j=\widetilde T_{kj}\exp(-\alpha)\exp(-\eta g_{kj}).
    \]
    Writing \(\eta=\gamma/\varepsilon\) and choosing \(\alpha\) to enforce
    \(\sum_jq_j=m_k\) gives exactly the tilted form above. Strict convexity of the KL
    objective on the positive orthant gives uniqueness.
\end{proof}

\begin{corollary}[Compatibility of row-wise bounds]
For a row with total mass \(m_k\geq0\), the attainable values of $\sum_{j=1}^{\ell} g_n(z_k,y_j)T_{kj}$ are exactly
$[m_k\min_j g_n(z_k,y_j),
        m_k\max_j g_n(z_k,y_j)]$. Therefore a target interval \([a_k,b_k]\) is feasible for row \(k\) if and only if $
    [a_k,b_k]
    \cap
    [
        m_k\min_j g_n(z_k,y_j),
        m_k\max_j g_n(z_k,y_j)
    ]
    \neq\varnothing .$
In particular, under the hard marginal constraint \(T\mathbf 1=\mu\), one has
\(m_k=\mu_k\), \(a_k=\mu_kd_-^n(z_k)\), and \(b_k=\mu_kd_+^n(z_k)\). Where $m_k = 0$, the attainable set reduces to $\{0\}$ and the same feasibility condition applies.

\label{cor:row-bounds}
\end{corollary}

\begin{proof}[Proof of Corollary \ref{cor:row-bounds}] 
Fix a row \(k\) and write
$g_{kj}:=g^n(z_k,y_j)$, 
    $g_k^{\min}:=\min_{1\leq j\leq \ell} g_{kj}$,
    $g_k^{\max}:=\max_{1\leq j\leq \ell} g_{kj}$.
Let \(T_{k\cdot}\in\mathbb R^\ell_+\) be any nonnegative row with total mass $\sum_{j=1}^{\ell}T_{kj}=m_k$.
Then
\[
    g_k^{\min} T_{kj}
    \leq
    g_{kj}T_{kj}
    \leq
    g_k^{\max}T_{kj}
    \qquad\text{for every }j.
\]
Summing over \(j\) gives
\[
    m_k g_k^{\min}
    \leq
    \sum_{j=1}^{\ell} g_{kj}T_{kj}
    \leq
    m_k g_k^{\max}.
\]
Thus every feasible row moment belongs to
\[
    [m_k g_k^{\min},\,m_k g_k^{\max}].
\]

Conversely, let
\[
    c\in [m_k g_k^{\min},\,m_k g_k^{\max}].
\]
If \(g_k^{\min}=g_k^{\max}\), then all \(g_{kj}\) are equal and the only attainable
moment is \(m_k g_k^{\min}=m_k g_k^{\max}\), so the claim is immediate. Otherwise,
choose indices \(j_-\) and \(j_+\) such that
\[
    g_{kj_-}=g_k^{\min},
    \qquad
    g_{kj_+}=g_k^{\max}.
\]
There exists \(\theta\in[0,1]\) such that
\[
    c
    =
    m_k\bigl(\theta g_k^{\min}+(1-\theta)g_k^{\max}\bigr).
\]
Define a row \(T_{k\cdot}\) by
\[
    T_{kj_-}:=m_k\theta,\qquad
    T_{kj_+}:=m_k(1-\theta),
\]
with all other entries equal to zero. Then
\[
    \sum_{j=1}^{\ell}T_{kj}=m_k
\]
and
\[
    \sum_{j=1}^{\ell}g_{kj}T_{kj}
    =
    m_k\bigl(\theta g_k^{\min}+(1-\theta)g_k^{\max}\bigr)
    =
    c.
\]
Hence the set of attainable row moments is exactly
\[
    [m_k g_k^{\min},\,m_k g_k^{\max}].
\]

It follows that a target interval \([a_k,b_k]\) is feasible for row \(k\) if and only if
\[
    [a_k,b_k]
    \cap
    [m_k g_k^{\min},\,m_k g_k^{\max}]
    \neq \varnothing .
\]
Indeed, feasibility requires the row moment to lie both in the prescribed target
interval and in the attainable moment interval, and the converse follows from the
attainability argument above.

In the hard-marginal case \(T\mathbf 1=\mu\), the row mass is \(m_k=\mu_k\).
Taking
\[
    a_k=\mu_k d_-^n(z_k),
    \qquad
    b_k=\mu_k d_+^n(z_k),
\]
gives the stated specialisation.
\end{proof}


\begin{corollary}[Complementary slackness for row-wise drift bounds]
Fix a row \(k\) with mass \(m_k>0\), and define $\psi_k(\gamma)$ as in Lemma \ref{lem:exponential-tilt}. Then \(\psi_k:\mathbb R\to\mathbb R\) is continuous and strictly decreasing. 
Let $([a_k,b_k])$ be a feasible target interval as in Corollary~\ref{cor:row-bounds}. Then the constrained projection onto this interval is characterised as follows:

\begin{itemize}
    \item If
\(\psi_k(0) < a_k,\)
then the lower bound is enforced by a unique \(\gamma_k^\star<0\) satisfying
\(\psi_k(\gamma_k^\star)=a_k.
\)
    \item If
\(
a_k \leq \psi_k(0) \leq b_k,
\)
then the untilted row already satisfies the admissibility interval, and no row tilt is required: \(\gamma_k^\star = 0\).
\item If \(\psi_k(0) > b_k,\) then the upper bound is enforced by a unique \(\gamma_k^\star>0\) satisfying
\(\psi_k(\gamma_k^\star)=b_k.
\)
\end{itemize}

Therefore, the projection onto \(\mathcal{C}_k\) reduces to a KL projection
onto the intersection of two affine constraints, which is given in closed form
by a single exponential tilting step (Lemma~\ref{lem:exponential-tilt}).
\label{cor:slackness}
\end{corollary}

\subsection{Proofs}\label{app:proofs}

\begin{proof}[Proof of Lemma \ref{lem:relaxed-sinkhorn}]
The objective is strictly convex in $T$ and coercive due to the KL penalties on both $T$ and its marginals, hence admits a unique minimiser.

Introducing dual variables for the marginal penalties, the first-order optimality conditions yield
\[
\log \frac{T_{kj}}{K_{0,kj}}
+ \lambda \log \frac{(T\mathbf{1})_k}{\mu_k}
+ \lambda \log \frac{(T^\top \mathbf{1})_j}{\nu_j}
= 0.
\]

Rearranging gives
\[
T_{kj}
=
K_{0,kj}\,
u_k\, v_j,
\]
where
\[
u_k = \exp\!\left(-\lambda \log \frac{(T\mathbf{1})_k}{\mu_k}\right),
\quad
v_j = \exp\!\left(-\lambda \log \frac{(T^\top \mathbf{1})_j}{\nu_j}\right).
\]

Thus the solution has the stated scaling form.
\end{proof}

\begin{proof}[Proof of Theorem \ref{thm:convergence}]
\label{prf:convergence}
The objective of the relaxed problem is a sum of KL divergences, hence strictly convex in $T$. The marginal penalties relax only the marginal constraints. Consequently, the relaxed problem admits a unique minimiser whenever
$C_+\cap C_-$ is nonempty, even if the hard marginal-constrained feasible set
$C_\mu\cap C_\nu\cap C_+\cap C_-$
is empty.



If the row-wise predictable-drift constraints are temporarily omitted, the relaxed problem reduces to
the soft-marginal entropic OT subproblem
\[
\min_{T\ge 0}
\mathrm{KL}(T\|K_0)
+
\lambda\left\{
\mathrm{KL}(T\mathbf 1\|\mu)
+
\mathrm{KL}(T^\top\mathbf 1\|\nu)
\right\}.
\]
Its minimiser has the scaling form
\[
T=\operatorname{diag}(u)K_0\operatorname{diag}(v),
\]
as in unbalanced Sinkhorn scaling.

The row-wise predictable-drift constraints are then imposed through Bregman
projections onto \(C_+\) or \(C_-\) due to complementary slackness (Corollary \ref{cor:slackness}). When a row constraint is active, Lemma~\ref{lem:row-tilt}
shows that the corresponding projection is a one-dimensional exponential tilt of that row.



\paragraph{Convergence of the algorithm.}

Each update in the proposed algorithm corresponds to a KL (Bregman) projection onto one of the constraint sets.

Since these sets are convex and the KL divergence is a Bregman divergence, and due to the complementary slackness of the drift constraints, the alternating projection scheme is a series of Bregman projections. Standard results on Bregman projections (e.g., \cite{bauschke2000dykstras, benamou2015iterative}) ensure convergence to the unique minimiser $T^*$.

If, in addition, the variance lower bound 
\[
\operatorname{Var}_{\pi_k(\gamma)}(g_n(z_k,Y)) \ge \sigma_k^2 > 0
\]
holds uniformly, then the algorithm converges linearly:
\[
\infdiv{T^{(t)}}{T^*}
\le
\rho^t \, \infdiv{T^{(0)}}{T^*}
\]
for some $\rho \in (0,1)$ depending on $(\varepsilon, \lambda, \{\sigma_k\})$.
\end{proof}

\begin{proof}[Proof of Lemma \ref{lem:quadratic-lb}]
Fix a row $k$ and write $g_j := g_n(z_k,y_j)$. By Lemma~\ref{lem:exponential-tilt}, the tilted plan takes the form
\[
T_{kj}(\gamma)
=
m_k \frac{\tilde T_{kj} \exp\!\left(-\gamma g_{kj} / \varepsilon \right)}
{Z(\gamma)},
\quad
Z(\gamma)
=
\sum_r \tilde T_{kr} \exp\!\left(-\gamma g_{kr} / \varepsilon \right).
\]
Then
\[
\log\frac{T_{kj}(\gamma)}{\tilde T_{kj}}
=
-\frac{\gamma}{\varepsilon} g_{kj} + \log m_k - \log Z(\gamma),
\]
and hence
\[
\infdiv{T(\gamma)}{\tilde T}
=
-\frac{\gamma}{\varepsilon}{\psi_k}(\gamma) + m_k\left[\log m_k - \log Z(\gamma)\right],
\]
where 
\[
	\psi_k(\gamma):=\sum_j g_jT_{kj}(\gamma).
	\]

Differentiating with respect to $\gamma$ yields
\[
\frac{d}{d\gamma}\text{KL}
=
-\frac{1}{\varepsilon}\psi_{k}(\gamma)
-\frac{\gamma}{\varepsilon}\psi'_{k}(\gamma)
- \frac{Z'(\gamma)}{Z(\gamma)}.
\]
A direct computation gives
\[
\frac{Z'(\gamma)}{Z(\gamma)}
=
-\frac{1}{m_k\varepsilon}\psi_{k}(\gamma),
\]
so the first and last terms cancel, leaving 
\[
\frac{d}{d\gamma}\text{KL}
=
-\frac{\gamma}{\varepsilon}\psi'_{k}(\gamma).
\]
By Lemma~\ref{lem:tilt}, we have
\[
\psi'_{k}(\gamma)
=
-\frac{m_k}{\varepsilon}
\operatorname{Var}_{\pi_{k}(\gamma)}(g_n(z_k,Y)),
\]
and therefore
\[
\frac{d}{d\gamma}\infdiv{T(\gamma)}{\tilde T}
=
\frac{{m_k}\gamma}{\varepsilon^2}
\operatorname{Var}_{\pi_{k}(\gamma)}(g_n(z_k,Y)).
\]
Since {\(T_k(0)=\widetilde T_k\), we have} $\infdiv{T_{k}(0)}{\tilde T_{k}}=0$. Integrating from $0$ to $\gamma$ {then yields}
\[
\infdiv{T_{k}(\gamma)}{\tilde T_{k}}
=
\frac{{m_k}}{\varepsilon^2} \int_0^\gamma  \operatorname{Var}_{\pi_{k}(s)}(g_n(z_k,Y)) \ s \, ds.
\]
Applying the uniform lower bound $\operatorname{Var}_{\pi_k(\gamma)}(g_n(z_k,Y)) \ge \sigma_k^2$ from Lemma~\ref{lem:tilt} yields
\[
\infdiv{T_{k}(\gamma)}{\tilde T_{k}}
\ge
\frac{{m_k}}{\varepsilon^2} \int_0^\gamma 
{s }\, ds
=
\frac{{m_k} \ \sigma_k^2}{2\varepsilon^2}\gamma^2.
\]
\end{proof}


\section{Computational complexity of the Bregman-Dykstra scheme}\label{app:complexity}
We compare the proposed Bregman--Dykstra implementation with a generic convex-optimisation baseline implemented in \texttt{cvxpy}, both solving \eqref{eq:Tref-Tcon} for the drift-constrained $T^\mathrm{proj}$. For each problem size, both methods solve the same finite-dimensional projection problem, with the same cost matrix, entropic regularisation parameter, marginal penalty, and constraints.

\begin{figure*}[ht]
    \centering
    \begin{subfigure}[t]{0.4\linewidth}
        \centering
        \includegraphics[width=\textwidth]{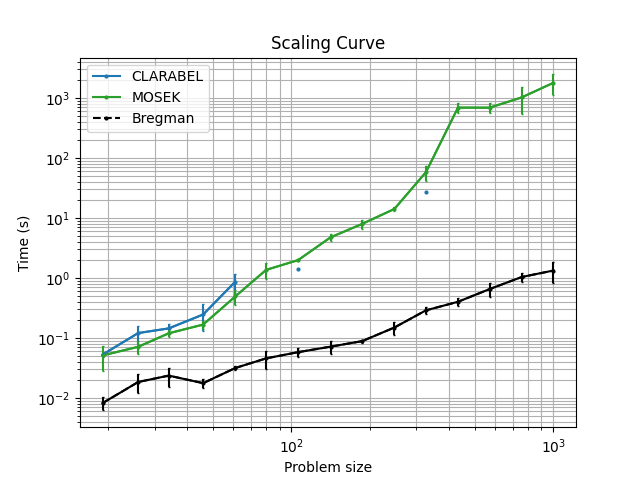}
        \caption{Scaling curves.}
    \end{subfigure}%
    ~ 
    \begin{subfigure}[t]{0.4\linewidth}
        \centering
        \includegraphics[width=\textwidth]{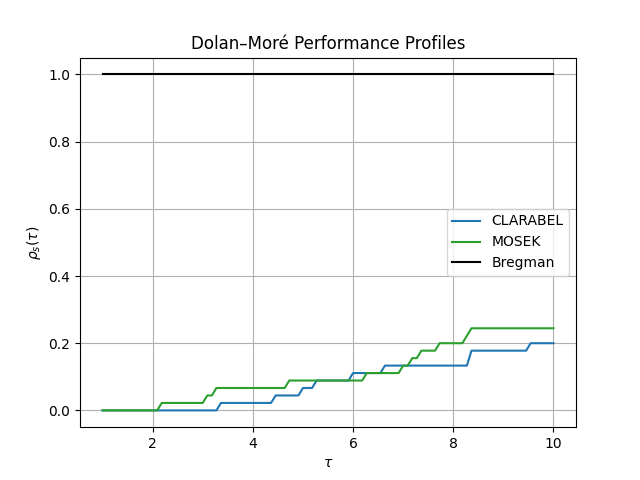}
        \caption{Dolan-Mor\'e performance profiles.}
    \end{subfigure}
    \caption{Computational comparison between the proposed Bregman--Dykstra projection algorithm and a generic \texttt{cvxpy} solver. Panel (a) reports scaling curves as the discretisation resolution increases. Panel (b) reports Dolan--Mor\'e performance profiles, showing the fraction of instances solved within a given runtime factor of the fastest method.}
    \label{fig:scaling-cvxpy}
\end{figure*}

Figure~\ref{fig:scaling-cvxpy} shows the runtimes, in seconds, on an Apple M2 Pro (16 GB) machine running MacOS Tahoe 26.4.1. No parallelisation or warm-starting is used in either setting. The specialized algorithm uses the structure of the entropic projection problem: marginal updates reduce to
Sinkhorn-type scalings, while predictable-drift constraints reduce to row-wise one-dimensional dual updates. This structure is not exploited by a generic solver.
The resulting scaling behaviour supports the use of the Bregman--Dykstra formulation as a practical implementation of the diagnostic.

\end{document}